%% file: root.tex
\documentclass[letterpaper, 10 pt, conference]{ieeeconf}  

\IEEEoverridecommandlockouts                              

\usepackage{graphics} 
\usepackage{mathptmx} 
\usepackage{cite}
\usepackage{amsmath}
\usepackage{graphicx}
\usepackage{float}
\usepackage{xcolor}
\usepackage{flushend}
\usepackage{array}
\usepackage{blindtext}
\usepackage{url}
\usepackage[binary-units = true]{siunitx} 
\usepackage[normalem]{ulem}
\usepackage{comment}
\usepackage{subcaption}

\newcommand{\lincolnbeet}{LB\xspace} 
\newcommand{\belgiumbeet}{BB\xspace} 

\title{\LARGE \bf
Towards practical object detection for weed spraying in precision agriculture*
}

\author{Adrian Salazar Gomez$^{1}$ \and Madeleine Darbyshire$^{2}$ \and Junfeng Gao$^{1}$ \and Elizabeth I Sklar$^{1}$ \and Simon Parsons$^{2}$
\thanks{*This work was partially supported by Lincoln Agri-Robotics as part of the Expanding Excellence in England (E3) Programme and by Ceres under the ``AI Unleashed'' project.}
\thanks{$^{1}$ASG, JG and ES are with the Lincoln Institute of Agri-food Technology, University of Lincoln, UK.
        {\tt $\{$asalazargomez, jugao, esklar$\}$@lincoln.ac.uk}}%
\thanks{$^{2}$MD and SP are with the Lincoln Centre for Autonomous Systems, University of Lincoln, UK.
        {\tt 25696989@students.lincoln.ac.uk, sparsons@lincoln.ac.uk}}%
\thanks{*This work has been submitted to the IEEE for possible publication. Copyright may be transferred without notice, after which this version may no longer be accessible}
        {\tt\small }}%
\begin{document}

\maketitle
\thispagestyle{empty}
\pagestyle{empty}

\begin{abstract}


The evolution of smaller, faster processors and cheaper digital storage mechanisms across the last 4-5 decades has vastly increased the opportunity to integrate intelligent technologies in a wide range of practical environments to address a broad spectrum of tasks.
One exciting application domain for such technologies is \emph{precision agriculture}, where the ability to integrate on-board machine vision with data-driven actuation means that farmers can make decisions about crop care and harvesting at the level of the individual plant rather than the whole field.
This makes sense both economically and environmentally.
However, the key driver for this capability is fast and robust machine vision---typically driven by machine learning (ML) solutions and dependent on accurate modelling.
One critical challenge is that the bulk of ML-based vision research  
considers only metrics that evaluate the accuracy of object detection and do not assess practical factors.
This paper introduces three metrics that highlight different aspects relevant for real-world deployment of precision weeding and demonstrates their utility through experimental results.
\end{abstract}
\begin{keywords}
precision agriculture, automated weeding, computer vision, object detection
\end{keywords}

\section{INTRODUCTION}
\label{sec:intro}

The current agricultural approach 
to weeding in arable crops is to spray an entire field with a selective herbicide that kills the weeds, but does not harm the crops.
Such a \emph{broadcast spraying} approach is easy to deliver---
requiring only a sprayer to dispense the herbicide---but wasteful, since much of the area 
sprayed does not contain weeds, being either bare earth or crops (see Figure~\ref{fig:experiment_fields}).
\emph{Precision agriculture} aims to use ideas from \emph{artificial intelligence~(AI)} and robotics to create agricultural solutions that can be delivered at the level of individual plants, instead of entire fields.
An important application within precision agriculture is \emph{automated weeding}, which aims to detect and target individual weeds, resulting in
the precise delivery of herbicide~\cite{Wufieldrobot2020} on the weeds while avoiding wastage, using a laser~\cite{MATHIASSEN2006497}, or a mechanical tool~\cite{Wufieldrobot2020}.
In the work presented here, we are concerned with precise delivery of herbicide in a real-world farm setting.

Now, it is clear that any approach to automated weeding needs to identify the weeds, and there have been numerous attempts to use computer vision to do this (see Section~\ref{sec:related}), many using AI methods based on \emph{machine learning~(ML)}.
However, most of this work has treated automated weeding as purely a computer vision problem: datasets are collected and annotated, object identifiers are trained, and the resulting models are optimised with respect to 
\emph{accuracy} and/or \emph{mAP} for classifying 
crops and weeds.
We argue that while these 
metrics are important, there are additional measures that must be considered in assessing the feasibility of ML models for use in precision spraying.
Here we focus on three of these.

\begin{figure*}[t]
\begin{minipage}[c]{.395\linewidth}
\begin{center}
\includegraphics[width=0.7\linewidth]{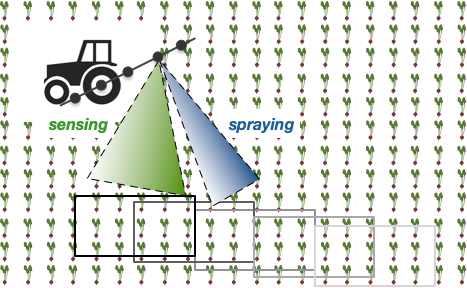}
\end{center}
\subcaption{}\label{fig:spray:a}
\end{minipage}
\begin{minipage}[c]{.595\linewidth}
\begin{center}
\includegraphics[width=0.85\linewidth]{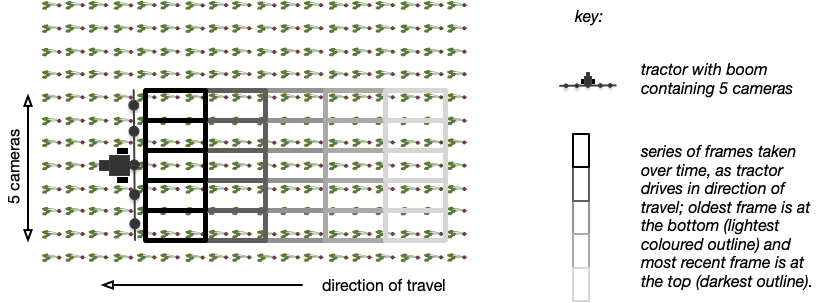}
\end{center}
\subcaption{}\label{fig:spray:b}
\end{minipage}
\caption{A diagrammatic explanation of a small precision sprayer. (a)  A boom is mounted on the back of a vehicle, carrying 5 cameras and a set of spray nozzles. Images are captured in real-time, as the vehicle moves. The system processes images, identifies weeds and their location within images, and targets spraying to hit the weeds.
Sequences of images captured may provide overlapping views of the field, or there may be gaps between images.
(b) A top-down view of the sprayer. The direction of travel for the is indicated in the drawing, from the bottom to the top of the illustration. Each camera captures its own image stream over time. Each row of images represents the set captured by the cameras at the same point in time; each column represents the passage of time as the tractor drives. The images here do not overlap.}
\label{fig:sprayer}
\end{figure*}

First, the \emph{weed coverage rate (WCR)} 
is more important than conventional ML-based metrics.
WCR identifies the proportion of the weeds that could be targeted by a sprayer triggered by the model.
WCR depends on the accuracy of the model, but also takes into account the resolution of a spray.
Second, we need to understand the \emph{area sprayed}, that is the area covered in herbicide by the precision sprayer in order to understand the saving in herbicide compared with current practice.
Third, it is important to know whether the approach is \emph{practical}---because you can't just pile more GPUs on-board a tractor operating in an open field, where issues like 
compute power and energy consumption come into play. Real-world 
automated weeding (see Figure~\ref{fig:sprayer}) will need to process many images very rapidly, so here we use \emph{inference speed} as a proxy for comparing the practicality of different ML approaches.

The primary contribution of this paper is to assess the feasibility of precision spraying by comparing 
a selection of standard ML-based 
vision methods applied to weed detection using the additional metrics we have introduced (WCR, area sprayed and inference speed).
Section~\ref{sec:related} briefly describes the state-of-the-art in ML-based vision approaches applied to agriculture.
Section~\ref{sec:methodology} explains our methodology and Section~\ref{sec:experiments} details the experiments we conducted which form the basis of our comparison.
Section~\ref{sec:discussion} analyses these results and Section~\ref{sec:conclusion} closes with conclusions.

\section{RELATED WORK}
\label{sec:related}

Initial approaches to weed detection used machine learning algorithms with handcrafted features based on the differences in colour, shape or texture. \cite{NGUYENTHANHLE2019116} extracted local binary patterns with support vector machines for plant discrimination. This method generally requires a relatively small dataset for model development.
However, it might fail to generalise under different field conditions. The deep learning-based methods for computer vision increasingly gain more popularity, offering an end-to-end weed detection solution that deals with the issue of generalisation. An adjusted YOLOv3 model was developed for bindweed detection in sugar beet fields \cite{gao2020deep}. This work provides good detection accuracy (\emph{m}AP=0.829), but it was based on the assumption of only one weed species in crop fields. \cite{jin2021deep} trained a CentreNet model only for crop detection and treated other remaining vegetation  objects as weed species. In this case, the model focuses on only crop detection without consideration the specific weed species in fields. Besides, \cite{Lottes_2018} developed a fully convolutional network integrating sequential information for robust weed detection in fields. The results show a better generalizability to unseen fields compared to other approaches in \cite{milioto2018realtime,lottes2017}. In order to remedy the scarcity of labelled samples for training the current deep supervised networks, \cite{Di_Cicco_2017} proposed a pipeline to generate annotated weed and crop synthetic data. 
Similar work based on the cut-and-paste approach for synthetic image generation can be seen in \cite{gao2020deep} for weed detection. Additionally, generative adversarial networks (GANs) were exploited in weed identification with transfer learning \cite{ESPEJOGARCIA202179}. 

Conventional CNNs require  millions of floating point parameters and consume a lot of compute resources.
Therefore, there have been many attempts to optimise networks for fast inference. 
An approach  that has been extremely successful is quantization. This aims to retain the performance of CNNs while decreasing the precision of weights and/or activations. Binarisation, a 1-bit quantization first used in \cite{courbariaux2016binaryconnect}, is the most extreme approach. BinaryNet\cite{courbariaux2016binarized} and XNOR-Nets\cite{rastegari2016xnornet} binarize both the weights and activations to achieve a further speed increase and a further reduction in memory usage. Bi-Real Net~\cite{liu2018BiRealNetECCV} aimed to improve the accuracy of these initial approaches by retaining the real-valued outputs of 1-bit convolution using a shortcut. BiDet~\cite{wang2020bidet} uses a binary network implementation based on Bi-Real Net for object detection using both SSD \cite{ssd} and Faster R-CNN~\cite{faster}. To our knowledge, there has been no other study investigating the use of binary neural networks for identifying weeds. In~\cite{corey2019binaryweeddetection}, nine varieties of weed in the DeepWeeds dataset were classified using a binary neural network. The authors found that most of the accuracy could be retained after downsampling the images to just 32x32x3 so running on the Intel DE1-SoC FPGA, an impressive inference time of 1.059ms could be achieved using just 6.04 watts of power. However, in this study we will focus on object detection instead of image classification and only look at how these networks perform on GPUs.  

After weed detection, following work on robotic weed control system can be studied for precision weed management. Spot spraying or selective spraying, the application case in this paper, is referred as switching the individual nozzle ON/OFF to deliver chemicals only for weed species. Hussain et al.~\cite{HUSSAIN2021106040,Hussain12244091} described a smart sprayer based on a deep learning detection model and evaluated the entire system under different weather conditions. Other than spraying, robotic mechanical precision weeding~\cite{Wufieldrobot2020} is an alternative, suitable for weed management in organic farming, but typically sacrificing the operation efficiency compared with spraying approach. Our study estimates the most suitable object detector based on deep learning by exploiting multiple datasets at different locations. This detector is determined with an emphasis on the balance between detection accuracy and inference speed for robotic spraying weeding development in field environments.  

\section{METHODOLOGY}
\label{sec:methodology}

The overall scenario we are considering is shown in Figure~\ref{fig:sprayer}.
A boom, mounted on the back of a farm vehicle, carries both cameras and nozzles for dispensing herbicide.
To be practical, images captured by the camera must be processed, and weeds detected, in time that they can be sprayed as the nozzles pass over them.
Here we describe how we evaluate the effectiveness of different approaches to this detection problem.

A conventional metric to evaluate the performance of object detectors is \textit{mean Average Precision} (\emph{m}AP).
To calculate mAP, first, a \emph{threshold} is defined for the \textit{IoU} value and this is used to distinguish between true positive ($TP$), false positive ($FP$) and false negative ($FN$) detections. The \emph{IoU} value, or \emph{Intersection over Union}, measures the accuracy of the predicted \emph{bounding box} circumscribing the object identified and is equal to $(A \cap B)/(A \cup B)$, where $A$ is the area of the ground truth (labelled) bounding box and $B$ is the area of the bounding box predicted by the model.

\emph{Precision} ($P$) is the proportion of correctly identified objects over all objects identified, $TP/(TP+FP)$.
\emph{Recall} ($R$) is the proportion of correctly identified objects over all identified objects (correct or incorrect), $TP/(TP+FN)$.
These metrics are used to compute \emph{average precision} ($AP$):
\[
AP = \sum_n( R_n - R_{n-1} ) . P_n
\]
where $n$ is the IoU threshold rank.
Because $AP$ is dependent on the IoU threshold value, mAP takes into account different threshold values and corresponding variations in the relationship between precision and recall.
Typically, $n=10$ threshold values are chosen, where the threesholds range from $0.5$ to $0.95$, i.e.  $\{0.5,0.55,0.6,\ldots,0.95\}$.

\input{images/examples_coverage_figure.tex}

While \emph{m}AP is a good measure of the performance of an object detector, it focuses on identifying the object very precisely.
In practice, for a sprayer, the level of precision for spraying is limited by the spray nozzle, and, as we will see, it is possible for an object detector that has a relatively low \emph{m}AP to still be good enough to be effective in ensuring that weeds are covered with herbicide.

To get a better idea of how the precision of the detector impacts the precision of spraying, we have devised a new metric which we call \emph{weed coverage rate (WCR)}. 
This estimates how many of the weeds in the test data would be sprayed, rather than how many are detected. 
We start by deciding how many sprayer nozzles $n$ will operate on the area within an image.
In practice it is possible to spray at a resolution of around 10cm which (see below) means that we could have 3 or 4 nozzles spraying independent sections of the ground covered by one image.
Each image $i$ is then split, along it length, into $n$ stripes.
If a weed is detected in a given stripe (that is the bounding box intersects with the stripe), an area that is the height of the stripe, and stretches to either side of the bounding box, is added to the spray area $S_i$.
This is illustrated in Figure~\ref{fig:weed:coverage}.

The spray area is larger than just the bounding boxes because the size is partly determined by the width of the spray (compare Figure~\ref{fig:weed:detection:1} and Figure~\ref{fig:weed:detection:3}). 
WCR differs from \emph{m}AP because that additional area, while increasing the herbicide used, will sometimes ``hit'' weeds that have evaded detection.
WCR is defined as follows.
A weed is counted as having been sprayed if it is wholly contained in the spray area:
\[
Sprayed(A_{i}) =
\begin{cases}
1, \text{if } A \subseteq S \\
0, \text{otherwise}
\end{cases}
\]
Then WCR is the fraction of the weeds that are counted as having been sprayed:
\[
WeedCoverageRate = \dfrac{\sum_i^n Sprayed(A_{i})}{n} \times 100
\]
In addition, we compare the area that has been sprayed with the total area of all the images:
\[
AreaSprayed = \dfrac{\sum_i^n S_{i}}{\sum_i^m I_{i}} \times 100
\]
Since area sprayed is a proportion of the total area that would be covered by broadcast spraying, 
The volume of herbicide saved is proportional to $100 - AreaSprayed$.

In order to establish the requirements in terms of frame-rate, we need to make some assumptions about the design of a sprayer.
The boom on a typical sprayer is 24m, and the recommended height to operate the boom above the crop canopy is 50cm \cite{boom:height}.
We established empirically that, at this height, a typical camera with a 1.8:1 aspect ratio (see below), can cover 550mm by 305mm.
The maximum speed of a sprayer is widely taken to be around 15mph, or 6.7m/s (see, for example,~\cite{sprayer:speed}) to prevent spray drift.
With the long edge of the image aligned along the spray boom, we need 44 cameras, and each camera can cover 305mm in the direction of travel. 
At 6.7m/s, we will need to process 22 frames per second (assuming no overlap), and across all 44 cameras the required frame-rate will be 968.
With the short edge of the image aligned along the boom we would need 79 cameras and process 13 frames a second per camera, giving a required frame-rate of 1027.

\section{EXPERIMENTS}
\label{sec:experiments}

\subsection{Datasets}

\begin{figure}[ht!]
\begin{center}
\includegraphics[width=0.95\columnwidth]{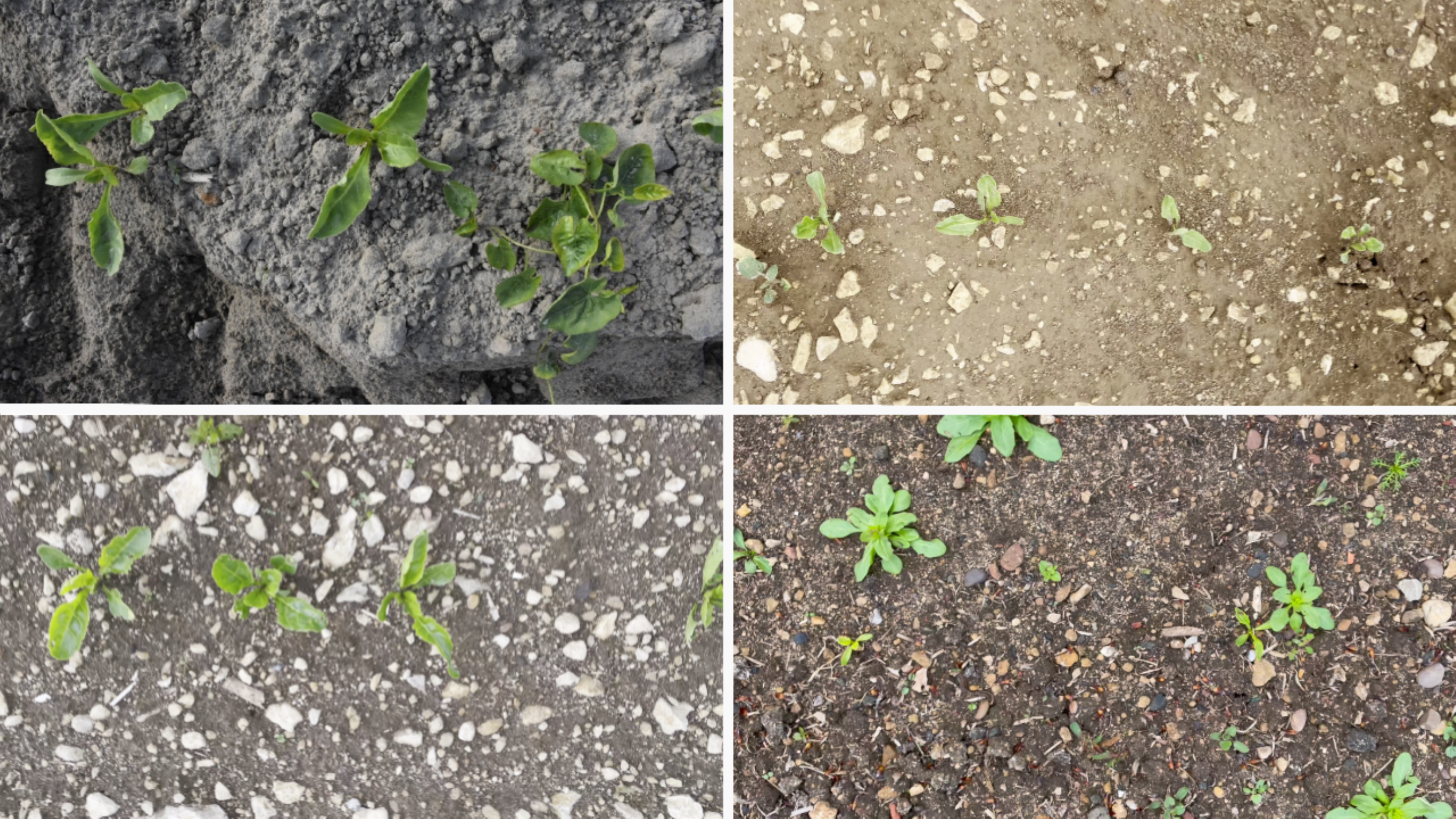}
\caption{Examples of the different fields in our datasets. Clockwise from top left: \belgiumbeet\ dataset, \lincolnbeet\ location 1, \lincolnbeet\ location 2, and \lincolnbeet\ location 3}
\label{fig:experiment_fields}
\end{center}
\end{figure}

 We used two weed/crop datasets: the \emph{Lincoln beet (\lincolnbeet)} dataset\footnote{The dataset will be available at the following link: \textbf{\url{https://github.com/LAR/lincolnbeet_dataset}}.} which we collected and annotated as part of this work and the \emph{Belgium beet (\belgiumbeet)} dataset from \cite{gao2020deep}. Both datasets contain pictures from fields in which sugar beet was grown commercially and the images contain pictures of beet and malicious weeds with their respective bounding box locations. The \belgiumbeet\ dataset contains 2506 images of $1800 \times 1200$ pixels. The \lincolnbeet\ dataset consists of 4405 images of $1920 \times 1080$ pixels.
The \lincolnbeet\ images were extracted from videos\footnote{The video frames converted to images were separated with enough frame distance to avoid repeated images in the dataset.} recorded at different points in time and three sugar beet fields. These data collection dates range from May to June 2021, where each field was scanned, at minimum, on four different dates with a week of separation to record weeds at different stages of growth. For all the scanning sessions, the distance from the camera to the ground was approximately 50cm. Two cameras were used; one with 12 megapixels, 26mm focal length and an f1.6 aperture and a camera with 64 megapixels, 29mm focal length and an aperture of f2.0. The original size of pictures from both cameras was $2160 \times 3840$ pixels. The fields used for the \lincolnbeet\ dataset are in Lincoln, UK, at different locations with conditions as the type of soil, distribution of the plants, and weed varieties. Fig. \ref{fig:experiment_fields} shows examples of the \belgiumbeet dataset and three fields used to create the \lincolnbeet\ dataset.
 
Both datasets present different item distribution and item visibility. The \belgiumbeet\ dataset has a lower number of items per picture than the \lincolnbeet\ dataset. In terms of visibility, the items in the \lincolnbeet\ dataset are proportionally smaller than the items in the \belgiumbeet\ dataset, and the items in the \belgiumbeet\ dataset have higher levels of inter-item occlusion. Table~\ref{tab:data_stats} shows the visibility and distribution characteristics of both datasets, and the characteristics of each of the items in the dataset. 

\input{tables/dataset_stats}

\subsection{Data preparation}

For the experiments, each dataset was randomly split into training, test and validation sets with 70\%, 20\%, and 10\% of the dataset images, respectively. To evaluate the approaches under different image resolutions, the images and the annotations were resized into three different sizes that maintain the width/height ratio; $960 \times 540$ pixels, $640 \times 360$ pixels, and $320 \times 180$ pixels. 

\subsection{Models and Parameters}
For the identification of weeds, we implemented one-stage detectors, two-stage detectors and a binary one-stage detector. The one-stage models are Yolov5m~\cite{yolov5}, Yolov3~\cite{yolov3} and Yolov4~\cite{yolov4}, where all three models use Darknet-53~\cite{yolov3} (DN-53) as a backbone. The two-stage detectors are based on Faster R-CNN~\cite{faster} with three different backbones: one with a ResNet-50 backbone~\cite{resnet} and a Feature Pyramid Network\cite{fpn} (FPN) neck; namely, ResNet-50-FPN (R-50-FPN), one with a ResNet-101-FPN~\cite{resnet} (ResNet-101-FPN), and another detector with a ResNeXt-101-FPN~\cite{restnext101} (Rx-101-FPN). During training, for the one-stage and two-stage models, the optimiser was stochastic gradient descent (SGD), the learning rate was \num{1e-2}, the momentum was \num{0.937}, and the learning decay was \num{5e-4}. In both model varieties, the networks were pre-trained on the COCO dataset\cite{coco}. The binary detector uses BiDet~\cite{wang2020bidet} and is based on the single shot detector (SSD)~\cite{ssd} with a Bi-Real Net~\cite{liu2018BiRealNetECCV} backbone based on VGG-16~\cite{vgg16}. The optimiser used for the binary model was ADAM~\cite{adam}; the learning rate was \num{1e-3}, momentum \num{0.9}, learning decay \num{0} and the network was pre-trained on Imagenet\cite{imagenet}.

Training and testing were run on a GTX2080Ti processor with 11 GB VRAM. For all detectors, the batch size, that is the number of images that are fed to the models simultaneously, used for training was the maximum number of images that can fit in the GPU along with the detector, and the number of epochs for training were 300 for the one-stage and two-stage detectors and 350 for the binary detector. For each model, the model weights used for testing were the ones with the highest \textit{m}AP on the validation set during the training process.  For testing and inference, the non-maximum-suppression threshold was $0.45$.

\subsection{Results}

We evaluated the trained models in several ways across both the \belgiumbeet and \lincolnbeet datasets and report both traditional metrics (\emph{m}AP) as well as our additional metrics: 
inference speed 
in \emph{frames per second (FPS)}, 
weed coverage rate (WCR) and 
area sprayed.
Analysis of results and discussion are given in the next section~(\ref{sec:discussion}).
Table~\ref{tab:standard_detectors_performance_lincoln} provides a conventional evaluation giving \emph{m}AP and speed of inference with batch size$=1$.
The approaches with the best performance on these metrics are highlighted.
Table~\ref{tab:batchsize_belgium} also provides inference speed, but exploits the ability of the models and the hardware they are run on to operate in parallel.
Results are given for a range of batch sizes.
Tables~\ref{tab:belgium_coverage} reports weed coverage rate and area sprayed for one to four nozzles. 
Figure~\ref{fig:nozzles_belgium} shows how WCR and spray area vary with the number of sprayer nozzles across the set of detectors. 
While weed coverage decreases with the number of nozzles, WCR decreases more slowly than area sprayed.
We also plotted the WCR and area sprayed against \emph{m}AP, Figure~\ref{fig:map:versus:metrics}.
Note that while we have results for all image sizes, space limitations mean we only show results for $640 \times 360$ images.

\input{tables/filtered_table_performance_belgium.tex}

\input{tables/filtered_multicamera_speed_belgium.tex}

\input{tables/filtered_coverage_belgium.tex}

\input{images/figure_nozzles_belgium.tex}

\input{images/map_vs_metrics.tex}

\section{DISCUSSION}
\label{sec:discussion}

Table~\ref{tab:standard_detectors_performance_lincoln} shows that  YoloV5m 
has the highest \emph{m}AP values
for both sugarbeet and weeds, on both the \lincolnbeet\ and \belgiumbeet\ datasets---a traditional ML evaluation would stop there.
But recalling our practical motivation to devise a strategy for assessing the feasibility of ML methods for precision spraying applications, we need to examine the results in more depth.
Table~\ref{tab:standard_detectors_performance_lincoln} shows that YoloV3 has the fastest inference speed.
The trade-off here is to select YoloV5m and process 68.96 frames per second very accurately or select YoloV3 and process 75.18 fps, but less accurately.

Next, we look at the trade-offs that come with different batch sizes.
Table~\ref{tab:standard_detectors_performance_lincoln} only reports results where batch size$=1$, whereas
table~\ref{tab:batchsize_belgium} compares results for higher batch sizes.
If images from 44 cameras were processed in parallel (i.e. batchsize$=44$) a frame rate of 277 or 333 (\belgiumbeet or \lincolnbeet, respectively) could be achieved with YoloV5m on a single GPU. 
While this is below the speed 
required to spray at our target pace of 15mph\footnote{As per Section~\ref{sec:methodology}, that would be a frame rate of 968 or 1027, depending on the number of cameras.}, a total of 3--4 GPUs would make in-the-field 
spraying with YoloV5m 
feasible.
 
Further, we investigate investigate the trade-offs 
when considering our WCR and area spray measures.
Figure~\ref{fig:nozzles_belgium} and table \ref{tab:belgium_coverage} contrast results when varying the number of sprayer nozzles per image from~1 to~4. 
For both datasets, using 3 nozzles seems to produce a significant reduction in area sprayed while retaining a good weed coverage rate---which is what we want: high WCR combined with low area sprayed means that we are hitting large numbers of weeds while wasting less herbicide.
While the highest WCR on the \belgiumbeet dataset is achieved by YoloV5m at $88.2$ for 3 nozzles, Faster-RCNN (50-FPN) only exhibits a small reduction in WCR with $85.78$ but only sprays $17.08\%$ of the area compared to $33.03\%$. 
For the \lincolnbeet dataset with 3 nozzles, 
YoloV5m achieves a WCR of $98.9$, but $68.6\%$ of the area is sprayed, whereas SSD achieves WCR of $96.94$ but only needs to spray $35.36\%$ of the area. 

Figure~\ref{fig:map:versus:metrics} illustrates the relationships between the traditional \emph{m}AP results (from table~\ref{tab:standard_detectors_performance_lincoln}) and the new WCR and area sprayed metrics.
While WCR and area sprayed are both positively correlated with \emph{m}AP, these results suggest that the detector which exhibits the ideal balance between weed coverage and herbicide usage may not be the detector with the highest \emph{m}AP.

One limitation of our results 
is that they were 
computed on non-contiguous images. This means the entire spray resolution needs to fit within 
one image. However, if our images were contiguous we could calculate the impact on 
WCR and area by allowing the spray to run over into the next image frame. Other limitations are that our model does not account for the time taken for the sprayer to turn on and off, and it assumes rectangular spray zones, whereas these are typically circular.
Spray drift due to wind force and vehicle movement is also an influencing factor for precision targeting. Modelling these aspects will provide more accurate estimates of sprayer performance, and they are all elements that we will address in future work.

\section{CONCLUSION}
\label{sec:conclusion}

This paper has demonstrated that selective spraying is feasible with state-of-the-art object detection. 
Our experiments show that the models considered are fast enough to detect and spray weeds on-the-fly, 
while our WCR metric helps to demonstrate, more clearly than conventional 
ML metrics, that the accuracy of our weed detection would be adequate to achieve a similar hit rate to broadcast spraying. At the same time, the area sprayed metric highlight options that produce 
clear reductions in area sprayed and hence herbicide required. 
As with many multi-criteria optimisation problems, there is no single clear winner.
Our new metrics help highlight the advantages and drawbacks of different approaches, demonstrating that when it comes to practical deployment, it's not just about \emph{m}AP.

In future, we will look to improve the frame rate achievable on a single GPU. Additionally, we will improve our WCR and area sprayed estimations by using contiguous images and take into account more properties of the spray, like area shape and sprayer response times. 

\clearpage
\bibliographystyle{IEEEtran}
\bibliography{IEEEabrv,references}

\end{document}

%% file: images/examples_coverage_figure.tex
\begin{figure*}[!ht]
\subcaptionbox{One nozzle\label{fig:weed:detection:1}}
{\includegraphics[width=0.24\textwidth]{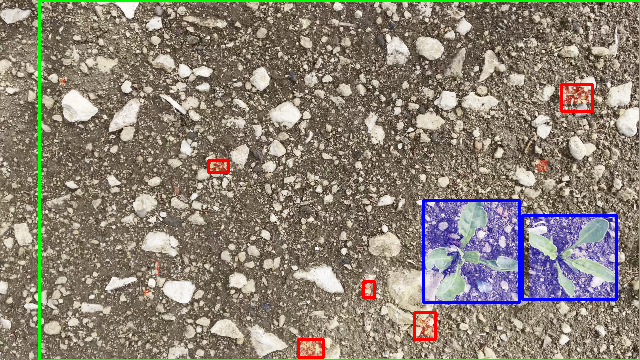}}
\subcaptionbox{Two nozzles\label{fig:weed:detection:2}}
{\includegraphics[width=0.24\textwidth]{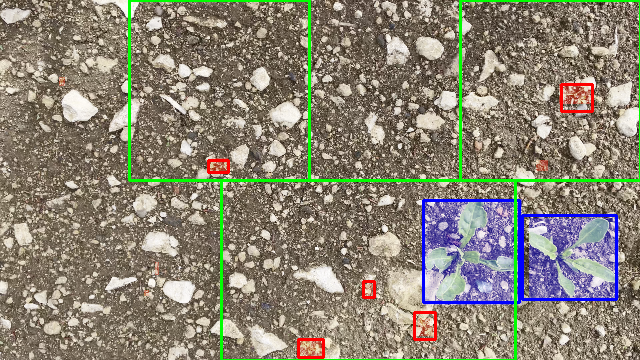}}
\subcaptionbox{Three nozzles\label{fig:weed:detection:3}}
{\includegraphics[width=0.24\textwidth]{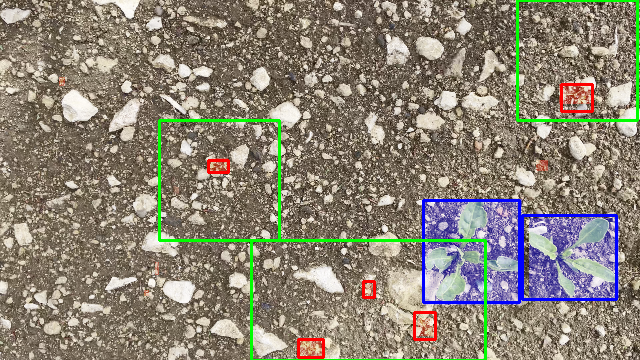}}
\subcaptionbox{Four nozzles\label{fig:weed:detection:4}}
{\includegraphics[width=0.24\textwidth]{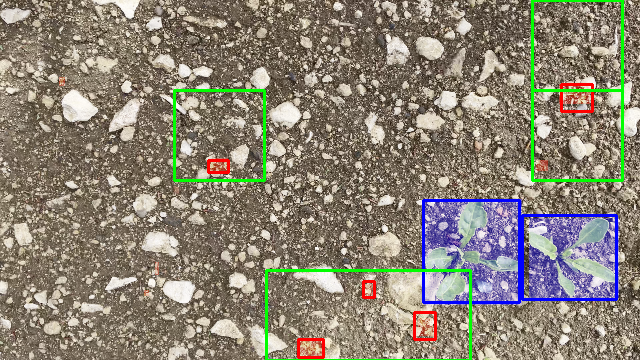}}
\caption{WCR. The spray area (green), weeds (red) and beet (blue) for the same image and one to four nozzles. 
When a nozzle is triggered, it sprays an area that starts before and finishes after (in the $x$ direction) the relevant weed bounding boxes and fills (in the $y$ direction) the relevant fraction of the image for the number of nozzles. Note how the area sprayed decreases as the number of nozzles increases, and how the rightmost weed triggers two ``sprays'' for 4 nozzles.}
\label{fig:weed:coverage}
\end{figure*}

%% file: tables/dataset_stats.tex

\begin{table}[!htb]
\centering
\scriptsize
\caption{Characteristics of the \belgiumbeet and \lincolnbeet datasets at dataset level(top) and at item level(bottom)}
\label{tab:data_stats}
\resizebox{\columnwidth}{!}{%
\begin{tabular}{|c|c|c|}
\hline
 &
  \belgiumbeet dataset & \lincolnbeet dataset \\ \hline
\begin{tabular}[c]{@{}c@{}}Number of images\end{tabular} & 2506       & 4402       \\ \hline
\begin{tabular}[c]{@{}c@{}}Number of items\end{tabular}  & 5578       & 39246      \\ \hline
\begin{tabular}[c]{@{}c@{}}Average items per picture\end{tabular} & 2.22  & 8.915   \\ \hline
\begin{tabular}[c]{@{}c@{}}Average percentage of the\\ bounding box\\  that is occluded\end{tabular} & 0.02187   & 0.0176     \\ \hline
\begin{tabular}[c]{@{}c@{}}Average area of the image \\ occupied by bounding boxes\end{tabular} & 0.09883 & 0.0717     \\ \hline
\begin{tabular}[c]{@{}c@{}}Average area of a bounding \\ box compared with the image area \end{tabular} & 0.0662 & 0.0272     \\ \hline
\begin{tabular}[c]{@{}c@{}}Average items \\ per pixel\end{tabular} & 2.5199e-06 & 4.2795e-06 \\ \hline
\multicolumn{3}{c}{}\\ \hline
 & \belgiumbeet dataset & \lincolnbeet dataset \\ \hline
\begin{tabular}[c]{@{}c@{}}Number of  sugar beets \end{tabular} & 2654  & 16399 \\ \hline
\begin{tabular}[c]{@{}c@{}}Average sugar beet plants per picture\end{tabular} & 1.059 & 3.725  \\ \hline
\begin{tabular}[c]{@{}c@{}}Number of weeds\end{tabular} & 2924 & 22847  \\ \hline
\begin{tabular}[c]{@{}c@{}}Average weed plants per picture\end{tabular} & 1.166 & 5.190  \\ \hline
\begin{tabular}[c]{@{}c@{}}Average area occluded in \\ sugar beet bounding boxes\end{tabular} & 0.0239 & 0.0324 \\ \hline
\begin{tabular}[c]{@{}c@{}}Average area occluded in \\  weed bounding boxes\end{tabular} & 0.0159 & 0.001  \\ \hline
\begin{tabular}[c]{@{}c@{}}Average image area occupied by \\ a sugar beet bounding box\end{tabular} & 0.0899 & 0.033  \\ \hline
\begin{tabular}[c]{@{}c@{}}Average image area occupied by \\ a weed\\ bounding box\end{tabular} & 0.0360 & 0.002  \\ \hline
\end{tabular}%
}
\end{table}

%% file: tables/filtered_table_performance_belgium.tex
\begin{table}[t]
\centering
\caption{Performance for object detectors with the \belgiumbeet (top) and \lincolnbeet (bottom) datasets. 
The fastest speeds (FPS) and 
highest \emph{m}AP scores are highlighted. 
}
\resizebox{\columnwidth}{!}{%
\begin{tabular}{|c|c|c|c|c|c|c|}
\hline
Model & Backbone  & FPS & \emph{m}AP Score & \begin{tabular}[c]{@{}c@{}} \emph{m}AP \\ Sugar Beet\end{tabular} & \begin{tabular}[c]{@{}c@{}}\textit{m}AP Score \\ Weed\end{tabular} \\ \hline
yolov5m                                                                  & DN-53    & 68.96  & \textbf{87.5} & \textbf{93.3} & \textbf{81.7}\\ \hline
Yolov4                                                                   & DN-53   & 46.29 & 63.2 & 74.8 & 51.5\\ \hline
Yolov3                                                                   & DN-53   & \textbf{75.18} & 85.6  & 92.1 & 79.1\\ \hline
Faster R-CNN                                                             & R-50-FPN    & 41.75  & 71.4 & 83.3 & 59.5\\ \hline
Faster R-CNN                                                             & ResNet-101-FPN   & 33  & 72.0  & 83.9  & 60.2\\ \hline
Faster R-CNN                                                             & Rx-101-FPN    & 18.2 & 75.7 & 85.8 & 65.6\\ \hline
SSD                                                                      & \begin{tabular}[c]{@{}c@{}}VGG-16 (Bi-Real)\end{tabular} & 58.01 & 63.9 & 75.6 & 52.2\\ \hline
\multicolumn{6}{c}{}\\ \hline
Model & Backbone &  FPS & \emph{m}AP Score & \begin{tabular}[c]{@{}c@{}}\\ \emph{m}AP \\ Sugar Beet\end{tabular} & \begin{tabular}[c]{@{}c@{}}\textit{m}AP Score \\ Weed\end{tabular} \\ \hline
yolov5m                                                                  & DN-53  & 49.26  & \textbf{51.0} & \textbf{67.5} & \textbf{34.6}\\ \hline
Yolov4                                                                   & DN-53  & 24.75 & 41.2 & 59.5 & 23.0\\ \hline
Yolov3                                                                   & DN-53  & \textbf{82.64} & 50.4  & 66.3 & 34.6\\ \hline
Faster R-CNN                                                             & R-50-FPN  & 40.13 & 42.4 & 62.2 &22.6 \\ \hline
Faster R-CNN                                                             & R-101-FPN  & 32.7 & 42.2 & 62.2 & 22.3\\ \hline
Faster R-CNN                                                             & Rx-101-FPN & 25.97 & 43.2 & 62.8 & 23.6\\ \hline
SSD & \begin{tabular}[c]{@{}c@{}}VGG-16 (Bi-Real)\end{tabular}  & 52.57 & 37.4 & 54 & 20.8\\ \hline
\end{tabular}
}
\label{tab:standard_detectors_performance_lincoln}
\end{table}

%% file: tables/filtered_multicamera_speed_belgium.tex
\begin{table}[t]
\centering
\caption{Speed in frames per second (FPS) for object detectors with different batch sizes for the \belgiumbeet dataset (top) and \lincolnbeet (bottom).}
\resizebox{\columnwidth}{!}{%
\begin{tabular}{|c|c|c|c|c|c|c|c|}
\hline
Model & Backbone & \begin{tabular}[c]{@{}c@{}}44\\ batch\end{tabular} & \begin{tabular}[c]{@{}c@{}}22\\ batch\end{tabular} & \begin{tabular}[c]{@{}c@{}}15\\ batch\end{tabular} & \begin{tabular}[c]{@{}c@{}}11\\ batch\end{tabular} & \begin{tabular}[c]{@{}c@{}}8\\ batch\end{tabular} & \begin{tabular}[c]{@{}c@{}}4\\ batch\end{tabular} \\ \hline
yolov5m                                                                  & DN-53  & 277 & 294.11 & 285 & 285 & 250 & 250\\ \hline
Yolov4                                                                   & DN-53  & 133 & 123 & 114 & 120 & 113 & 106\\ \hline
Yolov3                                                                   & DN-53  & 153 & 166 & 166 & 178 & 166 & 144\\ \hline
Faster R-CNN                                                             & R-50-FPN  & 58 & 55 & 55 & 55 & 50  & 50\\ \hline
Faster R-CNN                                                             & R-101-FPN  & 41 & 45.54 & 50.1 & 45.5 & 45.5 & 38\\ \hline
Faster R-CNN                                                             & Rx-101-FPN & 41 & 41 & 41 & 41 & 38 & 35 \\ \hline
SSD                                                                      &  \begin{tabular}[c]{@{}c@{}}VGG-16 (Bi-Real)\end{tabular}   & 62.8 & 66.4 & 70.3 & 67.7 & 63.9 & 61.2 \\ \hline
\multicolumn{8}{c}{}\\ \hline
%

Model & Backbone & \begin{tabular}[c]{@{}c@{}}44\\ batch\end{tabular} & \begin{tabular}[c]{@{}c@{}}22\\ batch\end{tabular} & \begin{tabular}[c]{@{}c@{}}15\\ batch\end{tabular} & \begin{tabular}[c]{@{}c@{}}11\\ batch\end{tabular} & \begin{tabular}[c]{@{}c@{}}8\\ batch\end{tabular} & \begin{tabular}[c]{@{}c@{}}4\\ batch\end{tabular} \\ \hline
yolov5m                                                                  & DN-53   & 333 & 333 & 344 & 312 & 312 & 208\\ \hline
Yolov4                                                                   & DN-53  & 40 & 39 & 40 & 38 & 39 & 38\\ \hline
Yolov3                                                                   & DN-53  & 188 & 175 & 175 & 181 & 188 & 169\\ \hline
Faster R-CNN                                                             & R-50-FPN  & 44 & 46 & 49 & 49 & 49  & 46\\ \hline
Faster R-CNN                                                             & R-101-FPN  & 38 & 41 & 44 & 41 & 41 & 40 \\ \hline
Faster R-CNN                                                             & Rx-101-FPN  & 35 & 36 & 36 & 36 & 35 & 33\\ \hline
SSD                                                                      & 
\begin{tabular}[c]{@{}c@{}}VGG-16 (Bi-Real)\end{tabular} & 25.9 & 25.8 & 26.1 & 25.1 & 23.5 & 21.5 \\ \hline
\end{tabular}
}
\label{tab:batchsize_belgium}
\end{table}

%% file: tables/filtered_coverage_belgium.tex
\begin{table*}[!ht]
\centering
\caption{WCR and area sprayed for 1--4 nozzles, for \belgiumbeet (top) and \lincolnbeet (bottom). Highlighted values explained in text.}
\label{tab:belgium_coverage}
\resizebox{0.98\textwidth}{!}{
\begin{tabular}{|c|c|c|c|c|c|c|c|c|c|}
\hline
      &          & \multicolumn{2}{c|}{1 nozzle} & \multicolumn{2}{c|}{2 nozzles} & \multicolumn{2}{c|}{3 nozzles} & \multicolumn{2}{c|}{4 nozzles} \\ \hline
Model & Backbone & \begin{tabular}[c]{@{}c@{}} Weed\\ coverage rate\end{tabular}     & \begin{tabular}[c]{@{}c@{}} Area \\ sprayed\end{tabular}     & \begin{tabular}[c]{@{}c@{}} Weed\\ coverage rate\end{tabular}       & \begin{tabular}[c]{@{}c@{}} Area \\ sprayed\end{tabular}   & \begin{tabular}[c]{@{}c@{}} Weed\\ coverage rate\end{tabular}       & \begin{tabular}[c]{@{}c@{}} Area \\ sprayed\end{tabular}     & \begin{tabular}[c]{@{}c@{}} Weed\\ coverage rate\end{tabular}       & \begin{tabular}[c]{@{}c@{}} Area \\ sprayed\end{tabular}     \\ \hline
yolov5m      & DN-53           & 100   & 84.21 & 96.75 & 48.46 & {\bf 88.82} & {\bf 33.03} & 78.73 & 25.97 \\ \hline
yolov4       & DN-53           & 100   & 83.52 & 96.43 & 46.88 & 87.68 & 31.51 & 76.05 & 24.67 \\ \hline
yolov3       & DN-53           & 100   & 84.51 & 95.45 & 48.34 & 86.89 & 33.03 & 79.06 & 25.9  \\ \hline
Faster R-CNN & R-50-FPN        & 99.03 & 67.31 & 94.52 & 28.34 & {\bf 85.78} & {\bf 17.08} & 75.08 & 12.51 \\ \hline
Faster R-CNN & R-101-FPN & 98.86 & 66.53 & 93.06 & 27.77 & 85.21 & 16.28 & 73.47 & 12.04 \\ \hline
Faster R-CNN & Rx-101-FPN & 99.68 & 67.61 & 93.87 & 28.18 & 84.57 & 16.67 & 73.79 & 12.11 \\ \hline
SSD          & \begin{tabular}[c]{@{}c@{}}VGG-16 (Bi-Real)\end{tabular} & 99.68 & 60.63 & 90.58 & 26.55 & 80.58 & 16.18 & 74.88 & 12.57 \\ \hline
\multicolumn{6}{c}{}\\ \hline
      &          & \multicolumn{2}{c|}{1 nozzle} & \multicolumn{2}{c|}{2 nozzles} & \multicolumn{2}{c|}{3 nozzles} & \multicolumn{2}{c|}{4 nozzles} \\ \hline
Model & Backbone & \begin{tabular}[c]{@{}c@{}} Weed\\ coverage rate\end{tabular}      & \begin{tabular}[c]{@{}c@{}} Area \\ sprayed\end{tabular}     & \begin{tabular}[c]{@{}c@{}} Weed\\ coverage rate\end{tabular} & \begin{tabular}[c]{@{}c@{}} Area \\ sprayed\end{tabular}     & \begin{tabular}[c]{@{}c@{}} Weed\\ coverage rate\end{tabular}       & \begin{tabular}[c]{@{}c@{}} Area \\ sprayed\end{tabular}     & \begin{tabular}[c]{@{}c@{}} Weed\\ coverage rate\end{tabular}       & \begin{tabular}[c]{@{}c@{}} Area \\ sprayed\end{tabular}     \\ \hline
yolov5m      & DN-53           & 99.1  & 99.4  & 99    & 87.2  & {\bf 98.9}  & {\bf 68.6}  & 98.7  & 55.3 \\ \hline
yolov4       & DN-53           & 99.1  & 97.3  & 97.6  & 73.6  & 95.2  & 50.8  & 92.9  & 39   \\ \hline
yolov3       & DN-53           & 99.1  & 99.5  & 99    & 87    & 98.8  & 68.3  & 98.5  & 55.2 \\ \hline
Faster R-CNN & R-50-FPN        & 98.5  & 84.6  & 95.9  & 50.4  & 92.8  & 31.7  & 88.7  & 22.2 \\ \hline
Faster R-CNN & R-101-FPN       & 98.4  & 83.7  & 96.2  & 49.9  & 93    & 31.2  & 88.8  & 21.5 \\ \hline
Faster R-CNN & Rx-101-FPN          & 98.2  & 81.1  & 94.9  & 46.9  & 91.6  & 29.2  & 86.9  & 20.2 \\ \hline
SSD          & \begin{tabular}[c]{@{}c@{}}VGG-16 (Bi-Real)\end{tabular} & 99.58 & 84.97 & 98.51 & 54.52 & {\bf 96.94} & {\bf 35.36} & 95.31 & 25.5 \\ \hline
\end{tabular}%
}
\end{table*}

%% file: images/figure_nozzles_belgium.tex
\begin{figure}[!ht]
\centering
\includegraphics[width=\columnwidth]{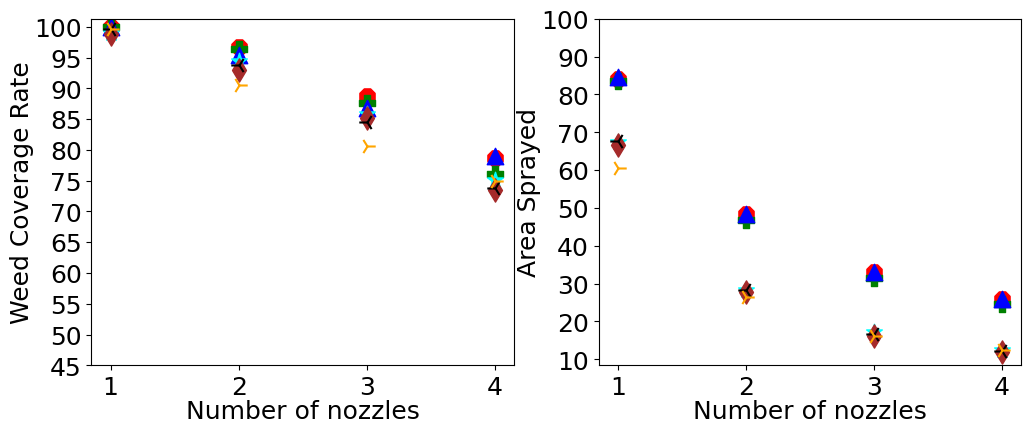}
\includegraphics[width=\columnwidth]{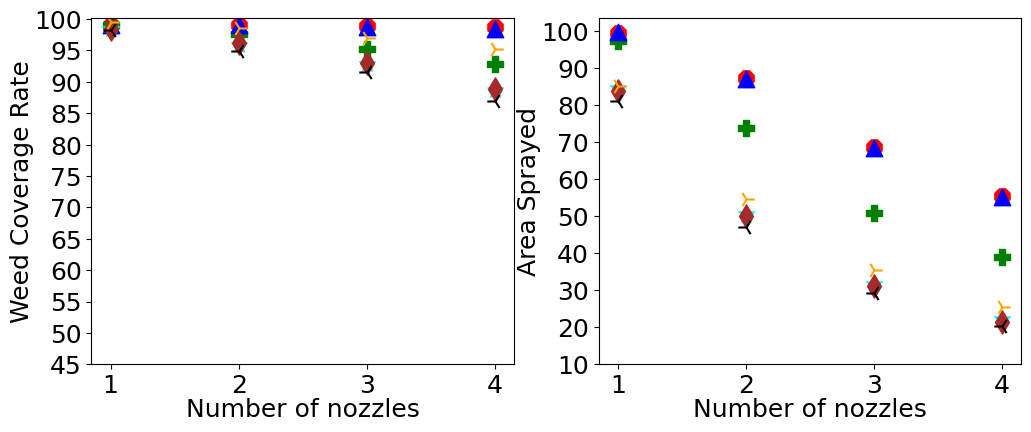}
\includegraphics[width=\columnwidth]{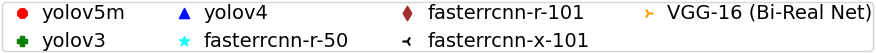}
\caption{The number of nozzles plotted against WCR (left) and area sprayed (right) for \belgiumbeet (top row) and \lincolnbeet (bottom).}
\label{fig:nozzles_belgium}
\end{figure}

%% file: images/map_vs_metrics.tex
\begin{figure}[!h]
\centering
\includegraphics[width=\columnwidth]{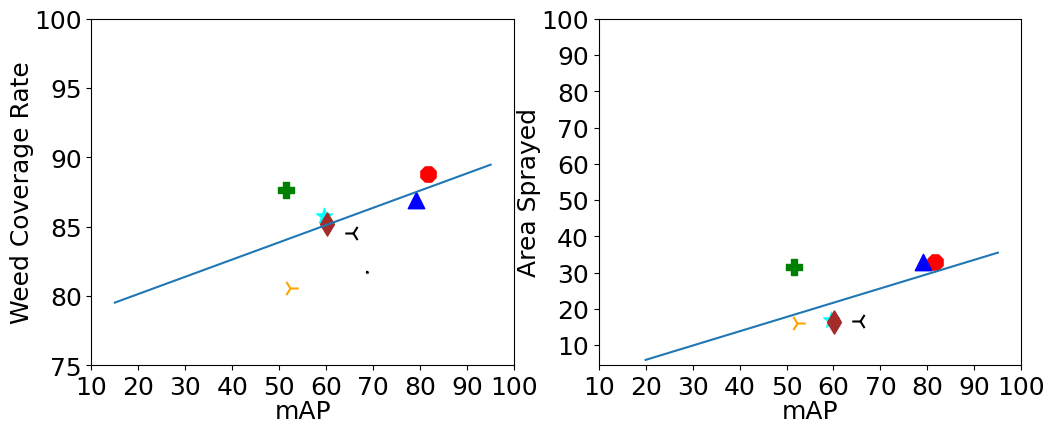}
\includegraphics[width=\columnwidth]{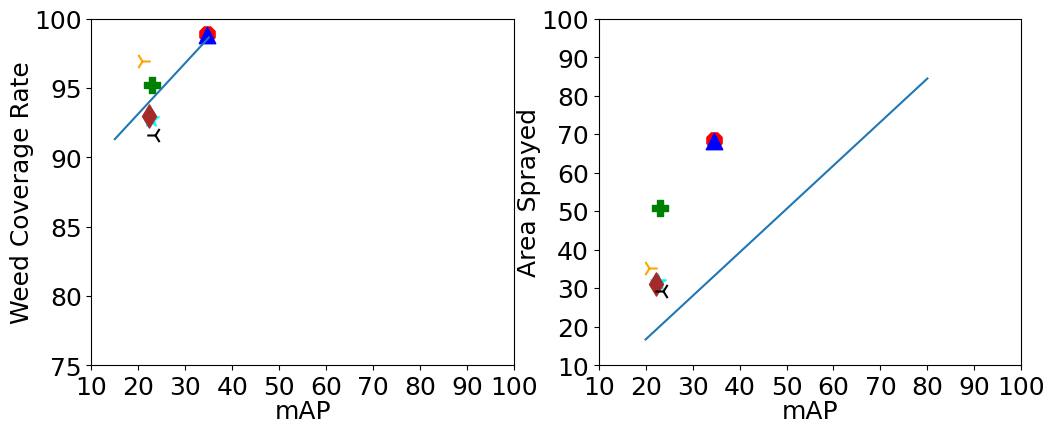}
\includegraphics[width=\columnwidth]{images/legend.png}
\caption{\textit{m}AP plotted against WCR (left) and area sprayed (right) with 3 nozzles for \belgiumbeet (top row) and \lincolnbeet (bottom).}
\label{fig:map:versus:metrics}
\end{figure}
